\documentclass{bmvc2k}

\title{Sparse and Privacy-enhanced Representation for Human Pose Estimation}

\addauthor{Ting-Ying Lin}{tingyinglin@m111.nthu.edu.tw}{1*}
\addauthor{Lin-Yung Hsieh}{linyunghsieh@gapp.nthu.edu.tw}{1*}
\addauthor{Fu-En Wang}{fulton84717@gapp.nthu.edu.tw}{1}
\addauthor{Wen-Shen Wuen}{vincent_wuen@novatek.com.tw}{2}
\addauthor{Min Sun}{sunmin@ee.nthu.edu.tw}{1}

\usepackage{bigstrut}
\usepackage{multirow}
\usepackage{graphicx}
\usepackage{float}
\usepackage{subfigure}
\usepackage{booktabs}
\usepackage{wrapfig}
\usepackage{amssymb}

\addinstitution{
 Vision Science Lab\\
 National Tsing Hua University\\
 Hsinchu, Taiwan
}
\addinstitution{
 Novatek Microelectronics Corp.\\
 Hsinchu, Taiwan
}

\runninghead{Lin et al.}{Sparse and Privacy-enhanced Representation for HPE}

\begin{document}

\maketitle
\definecolor{frenchblue}{rgb}{0.0, 0.45, 0.73}
\definecolor{gray}{rgb}{0.5,0.5,0.5} 
\definecolor{green}{rgb}{0, 0.4, 0} 
\definecolor{orange}{rgb}{1, 0.5, 0} 	
\definecolor{mahogany}{rgb}{0.75, 0.25, 0.0}
\definecolor{purple}{rgb}{0.6, 0, 0.6}
\definecolor{darkgreen}{rgb}{0, 0.4, 0.4} 
\definecolor{teal}{rgb}{0.0, 0.5, 0.5}
\definecolor{aaaa}{rgb}{0.55, 0.1, 0.7}
\definecolor{red}{rgb}{1.0, 0, 0}


\newcommand{\fuen}[1]{\textcolor{blue}{[FuEn]: #1}}
\newcommand{\leo}[1]{\textcolor{teal}{[leo]: #1}}
\newcommand{\tin}[1]{\textcolor{aaaa}{[tin]: #1}}
\newcommand{\minsun}[1]{\textcolor{mahogany}{[minsun]: #1}}
\vspace{-0.6\baselineskip}
\begin{abstract}
We propose a sparse and privacy-enhanced representation for Human Pose Estimation (HPE). Given a perspective camera, we use a proprietary motion vector sensor (MVS) to extract an edge image and a two-directional motion vector image at each time frame. Both edge and motion vector images are sparse and contain much less information (\textit{i.e.}, enhancing human privacy). We advocate that edge information is essential for HPE, and motion vectors complement edge information during fast movements. We propose a fusion network leveraging recent advances in sparse convolution used typically for 3D voxels to efficiently process our proposed sparse representation, which achieves about 13x speed-up and 96\% reduction in FLOPs. We collect an in-house edge and motion vector dataset with 16 types of actions by 40 users using the proprietary MVS. Our method outperforms individual modalities using only edge or motion vector images. We also demonstrate the generalizability of our approach on MMHPSD and HumanEva datasets. Finally, we validate the privacy-enhanced quality of our sparse representation through face recognition on CelebA (a large face dataset) and a user study on our in-house dataset. The code and dataset are available on the project page:~\url{https://lyhsieh.github.io/sphp/}.

\vspace{-0.6\baselineskip}
\end{abstract}


\section{Introduction}

With the advance of deep learning~\cite{chai2021deep} for visual perception and the lower cost of connected camera systems, human society is at the beginning of the smart camera systems era \cite{VIFeiFei} facilitating our daily life. However, there are two key challenges to be addressed. Firstly, the dilemma between convenience and privacy. We want the systems (\textit{e.g.}, in our home) to recognize our behavior and assist us, but we also need to ensure they protect our privacy. Secondly, applying deep models on the cloud's visual data is costly and introduces delays. We aim to run these models efficiently on the edge device, which supports real-time response and protects our privacy. Therefore, a sparse and privacy-enhanced representation for human pose estimation is desired.

An event-based camera is a sparse representation approach that only records the intensity variances between active pixels. Hence, privacy can be enhanced, and recorded identities cannot be directly recognized compared to traditional cameras. Human pose estimation from event data has been studied in recent years~\cite{dhp19,eventhpe}. However, networks trained with event data usually have a significantly lower performance in HPE due to no event data recorded on still body parts. In order to achieve better HPE performance by sparse representation, a new approach is necessary for privacy-enhanced smart cameras on edge devices. 


\begin{figure}
  \begin{minipage}[c]{0.55\textwidth}
    \includegraphics[width=\textwidth]{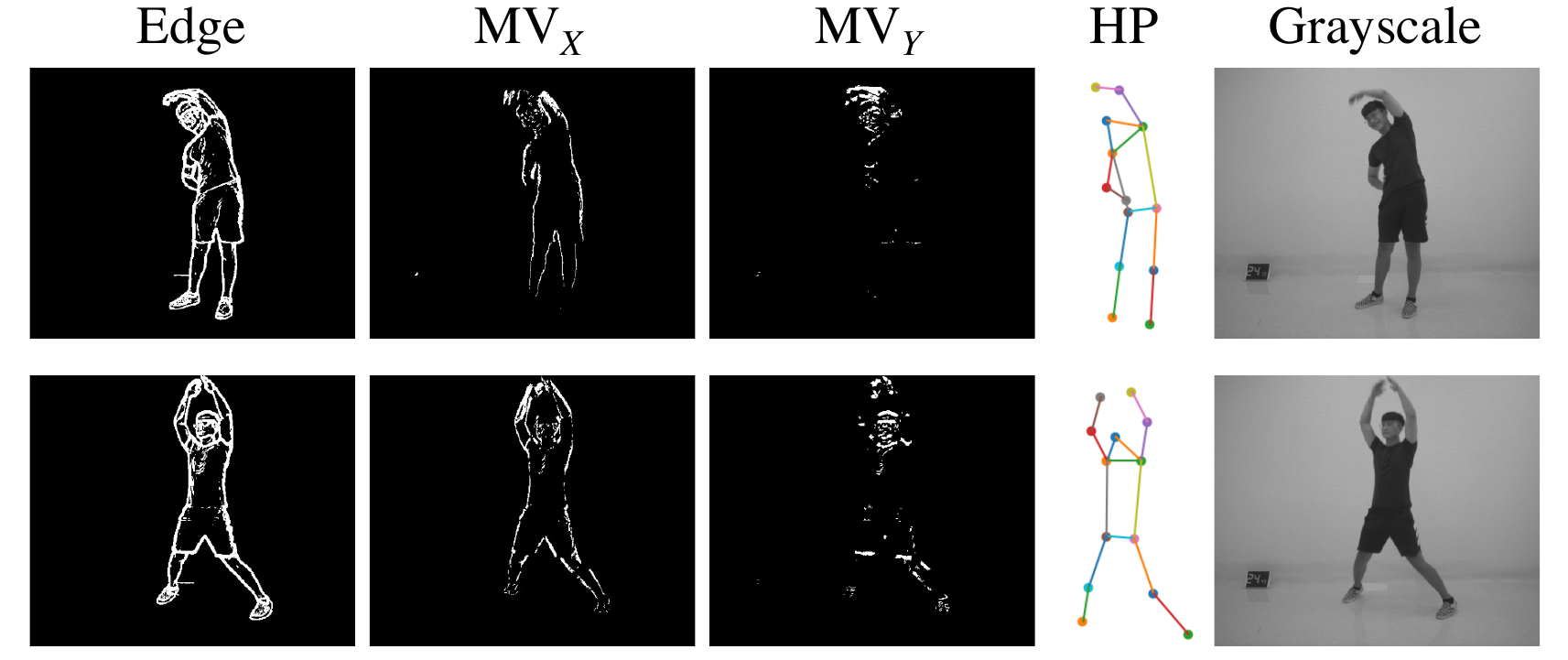}
  \end{minipage}\hfill
  \begin{minipage}[c]{0.45\textwidth}
    \caption{
    Examples from SPHP. Our motion vector sensor (MVS) can extract an edge image and a two-directional motion vector image (visualized as horizontal MV$_X$ and vertical MV$_Y$, respectively) at each time frame. We also provide annotated body joints for Human Poses (HP) and corresponding grayscale images.
    } \label{fig:dataset}
  \end{minipage}
  \vspace{-0.5\baselineskip}
\end{figure}


Inspired by the seminal edge (\textit{i.e.}, boundary)~\cite{Martin2004} and motion~\cite{Bro11a} literature before deep learning, we advocate that edge information is essential for HPE, and motion vectors complement edge information during fast movements.
Besides, both edge and motion vectors are sparse representations that can be efficiently processed and enhance privacy. Hence, we propose to use edge to capture the contour of all body parts and replace event data with two-directional motion vectors.
Specifically, we use a proprietary motion vector sensor (MVS) to extract an edge image and a two-directional motion vector image at each time frame. Thus, our method differs from pure software solutions that may already contain privacy-sensitive data.
We propose a fusion network leveraging these complementary sparse representations (\textit{i.e.}, edge and motion vectors) to perform better than each representation. Moreover, we exploit the sparsity of our data to replace standard convolution with submanifold sparse convolutions~\cite{scn} typically used for 3D voxels to speed up the inference time by 13 times.

Unlike RGB images, our proposed edge and motion vector representation lacks labeled data. Therefore, we collect an in-house dataset called Sparse and Privacy-enhanced Dataset for Human Pose Estimation (\textbf{SPHP}), as shown in Figure~\ref{fig:dataset}. This dataset includes recordings of 40 individuals performing 16 exercise and fitness-related actions, such as stretching or jogging, captured from 4 distinct viewing angles. 
We also collect grayscale images synchronized with edge and motion vectors to speed up joint position annotation. First, we apply a pre-trained natural-image-based HPE network on the grayscale images to obtain high-quality "initial" joint labels. Next, we ask human annotators to check and correct the errors if needed. This approach is estimated to save about 2,000 hours of manual joint labeling effort.

To demonstrate the applicability of our proposed framework, we begin by conducting extensive experiments on our \textbf{SPHP} dataset.
Our fusion method surpasses the performance of individual modalities (\textit{i.e.}, edge or motion vectors) for human pose estimation, especially for fast-moving joints, with a maximum relative improvement of 11\% over using only the edge modality. We also compare traditional and submanifold sparse convolutions to demonstrate the advantage of our sparse data. The results show that sparse convolution networks can reduce the number of FLOPs of HPE by 96\% while maintaining an acceptable error rate. In addition, we achieve a 13x acceleration in inference time, which is particularly beneficial for real-time applications. Besides, we confirm the generalizability of our approach on additional HPE datasets, such as MMHPSD\cite{eventhpe} and HumanEva\cite{humaneva}. The results on MMHPSD highlight that motion vectors capture more valuable information compared to traditional event cameras.
To verify the privacy-enhanced attribute of our data, we use ArcFace~\cite{arcface} to evaluate the cosine similarity between various modalities of faces on the large-scale face attributes dataset, CelebA~\cite{celebA}. Using edge as input results in a 10.2\% recall drop compared to RGB images, indicating that privacy is enhanced by converting RGB images to edge images. Moreover, we conduct a user study to check if humans can identify a specific individual from 10 leaked edge images, given a grayscale reference face. The significantly low accuracy of 18.8\% demonstrates the limited ability of humans to recognize faces in edge modality.

Our main contributions are summarized as follows:
\begin{enumerate}
    \item We introduce the Sparse and Privacy-enhanced Dataset for Human Pose Estimation (\textbf{SPHP}), which consists of synchronized, complementary images of edge and motion vectors along with ground truth labels for 13 joints.
    \item Our fusion method outperforms individual modalities (\textit{i.e.}, edge or motion vectors) for human pose estimation, particularly for fast-moving joints, with a maximum relative improvement of 11\% on our dataset when compared to using only the edge modality. Additionally, we further demonstrate the generalizability on other HPE datasets.
    \item The high sparsity of \textbf{SPHP} enables us to achieve a maximum 13x acceleration in inference time and decrease FLOPs by 96\% after applying sparse convolution, significantly reducing computational costs compared to traditional convolutional neural networks.
    \item We demonstrate the privacy-enhanced nature of our sparse representation through face recognition experiments. We utilize edge images as input and validate a recall drop of over 10.2\% in comparison to RGB images on the CelebA dataset. Furthermore, our user study shows that humans have limited ability to recognize faces in edge modality.
\end{enumerate}

\vspace{-0.8\baselineskip}

\section{Related work}

\subsection{Human Pose Estimation}
Human Pose Estimation (HPE) is a rapidly developing field of research in computer vision, intending to predict 2D/3D positions of joints from various types of input, including RGB, grayscale, or even depth image~\cite{marin2018hpedepth, shotton2011hpe, shotton2013hpe}. HPE has numerous potential applications, such as action recognition, health monitoring~\cite{cao2021inbed}, and physical education. 
Currently, CNN-based methods~\cite{fan2015dualsource, tain2012exploring, xu2022locationfree, yang2017learning, wei2016convolutional} achieve the state-of-the-art performance. Some focus on single-person pose estimation~\cite{alex2014deeppose, hourglass} while others can perform HPE on multi-person~\cite{cao2017realtime, fang2017rmpe, yang2023explicit}. These methods are typically trained and evaluated on RGB-based image datasets such as MS-COCO~\cite{mscoco}, MPII~\cite{MPII2014}, and CrowdPose~\cite{crowdpose}.

Top-down and bottom-up approaches are two common strategies in HPE. A top-down approach is a two-stage method that performs human detection at each input image to find the region of interest before predicting joint positions. The accuracy of a top-down approach may rely on the quality of the human detection process. Representative networks for top-down paradigms include HRNet~\cite{sun2019hrnet}, Mask-RCNN~\cite{maskrcnn}, CFN~\cite{huang2017cfn}, CPN~\cite{chen2018cpn}, G-RMI~\cite{papa2017grmi} and SimpleBaseline~\cite{xiao2018simplebaseline}.
A bottom-up approach detects each joint position and assembles them into groups. Representative works are DeepCut~\cite{deepcut}, DeeperCut~\cite{deepercut} and OpenPose~\cite{cao2017realtime}. Bottom-up approaches have shown greater efficiency in HPE for multiple humans compared to top-down approaches and can effectively handle complex poses involving self-occlusion. In this study, we adopt a bottom-up approach for our backbone networks.

\vspace{-0.5\baselineskip}

\subsection{Sparse Convolution}
Convolution Neural Networks (CNNs) have been proven effective for many visual recognition tasks, including human pose estimation. However, CNNs' high computational cost makes it challenging to use on resource-limited embedding systems. The cost significantly increases when higher dimensional convolutions, such as 3D CNN on 3D point clouds, are needed. Despite the challenge, several sparse convolution techniques have been proposed to leverage the sparsity property within the data. For instance,~\cite{ben2015sparse, martin2017vote3deep} 
propose sparse convolutional layers for processing 3D data efficiently.~\cite{scn} proposes submanifold sparse convolution, which leads to computational saving but remains state-of-the-art performance.~\cite{wang2017ocnn} designs a sparse data structure and realizes fast computation for 3D shape analysis.~\cite{hang2018splanet, gernot2017octnet} use sparse convolutional layers to maintain the efficiency for processing point clouds. Other than 3D data, handwritten characters are a 2D sparse data example demonstrating sparse convolution's effectiveness for character recognition~\cite{benjamin2014spatially}. Nevertheless, fewer studies~\cite{mathias2022deltacnn} have been conducted to leverage sparse convolution for human pose estimation.

\vspace{-0.5\baselineskip}

\subsection{Privacy Enhancing}
Using RGB images to recognize human behavior has many useful applications but raises privacy concerns. Hence, enhancing privacy using software or hardware-level techniques becomes an essential research direction. At the software level, early methods~\cite{padilla2015visual} involved algorithms such as encryption, image filtering, and object/people removal. Besides,~\cite{ryoo2017privacy} introduces inverse super-resolution, generating multiple low-resolution training videos (\textit{e.g.}, 16$\times$12) from high-resolution videos for human activity recognition.~\cite{wu2018privacy,ren2018learningprivacy} adopt adversarial learning to remove privacy-sensitive information from images while balancing the trade-off between privacy and task. At the hardware level, privacy-enhanced optics are designed to filter out private information while still retaining functionality such as human pose estimation and action recognition.~\cite{pittaluga2015privacy} applies privacy-enhanced optics to block sensitive information from the incident light field before sensor measurements.~\cite{arguello2022optics,hinojosa2022privhar,hinojosa2021learning} use optimal encoders to protect privacy, and use convolutional neural networks to extract features of specific tasks. In this work, we use a proprietary motion vector sensor (MVS) to extract an edge image and a two-directional motion vector image at the hardware level. We show that MVS significantly enhances privacy through our designed face recognition experiments.


\vspace{-0.6\baselineskip}
\section{Approach}
\vspace{-0.3\baselineskip}

We describe the detailed process to obtain our sparse representation (Section~\ref{sec.rep}). Then, we introduce a fusion model (Section~\ref{sec.fusion}), which benefits from both edge and motion vectors. To enhance the efficiency of our fusion model, we utilize sparse convolution (Section~\ref{sec.sparse-cov}).

\subsection{Sparse Representation}\label{sec.rep}

We use a proprietary Motion Vector Sensor (MVS) to extract an edge image and a two-directional motion vector image at each time frame.
Our MVS provides sparse and privacy-enhanced representation, as detailed below, compared to traditional RGB/grayscale cameras. 

\noindent\textbf{Edge Image.}
MVS uses an efficient hardware implementation of edge detection, similar to Canny edge detection~\cite{canny}, to generate edge images. Each pixel in the edge image has a value within the range of \{0, 255\}. A higher value indicates a stronger intensity of the edge.

\noindent\textbf{Motion Vector.}
Inspired by the motion detection of the Drosophila visual system~\cite{flyvision} and designed with patent-pending pixel-sensing technology, MVS detects vertical and horizontal motion vectors, denoted as MV$_X$ and MV$_Y$ shown in Figure~\ref{fig:dataset}, by analyzing changes in illumination. Each value falls within the range of \{-128, 128\}. The magnitude and sign of a value represent the strength and direction of motion, respectively.

Since MVS only produces non-zero values when detecting edge or motion vectors, the resulting data can be very sparse. This greatly reduces resource usage during computation and storage. Additionally, the output data from MVS only focuses on changes in illumination over time and space, reducing the risk of privacy exposure (\textit{i.e.}, privacy-enhanced).


\subsection{Fusion Model}\label{sec.fusion}

\begin{figure}[!htb]
    \centering
    \begin{minipage}{0.485\textwidth}
        \centering
\        \includegraphics[width=\textwidth]{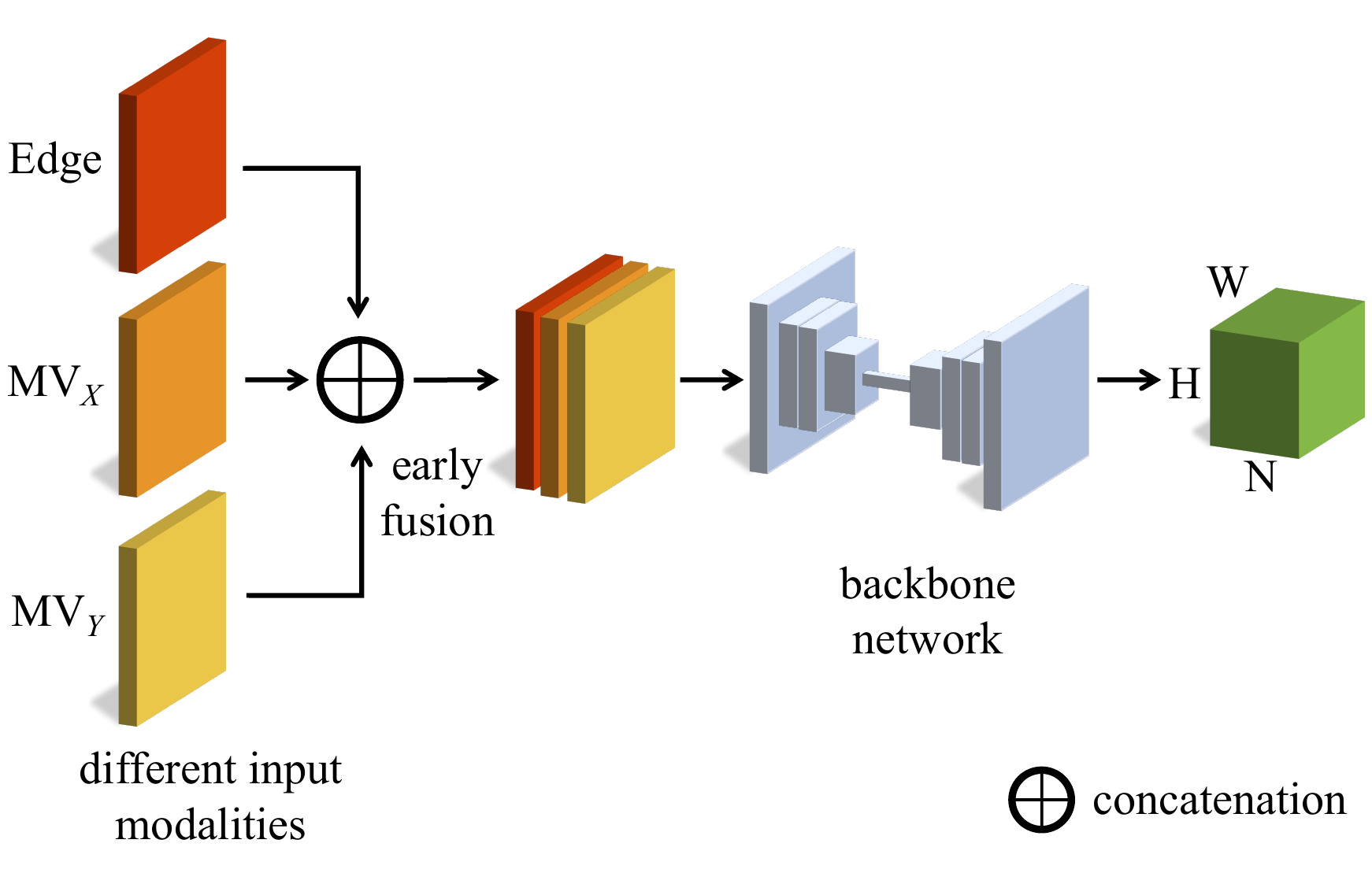}
        \vspace{-1.1\baselineskip}
        \caption{Fusion model. We employ early fusion on different data modalities, including edge and two-directional motion vectors (MV$_X$ and MV$_Y$). The resulting output channel N corresponds to the number of joints.}
        \label{fig:fusion}
    \end{minipage}
    \begin{minipage}{0.03\textwidth}
    \end{minipage}
    \begin{minipage}{0.43\textwidth}
        \centering
        \includegraphics[width=\textwidth]{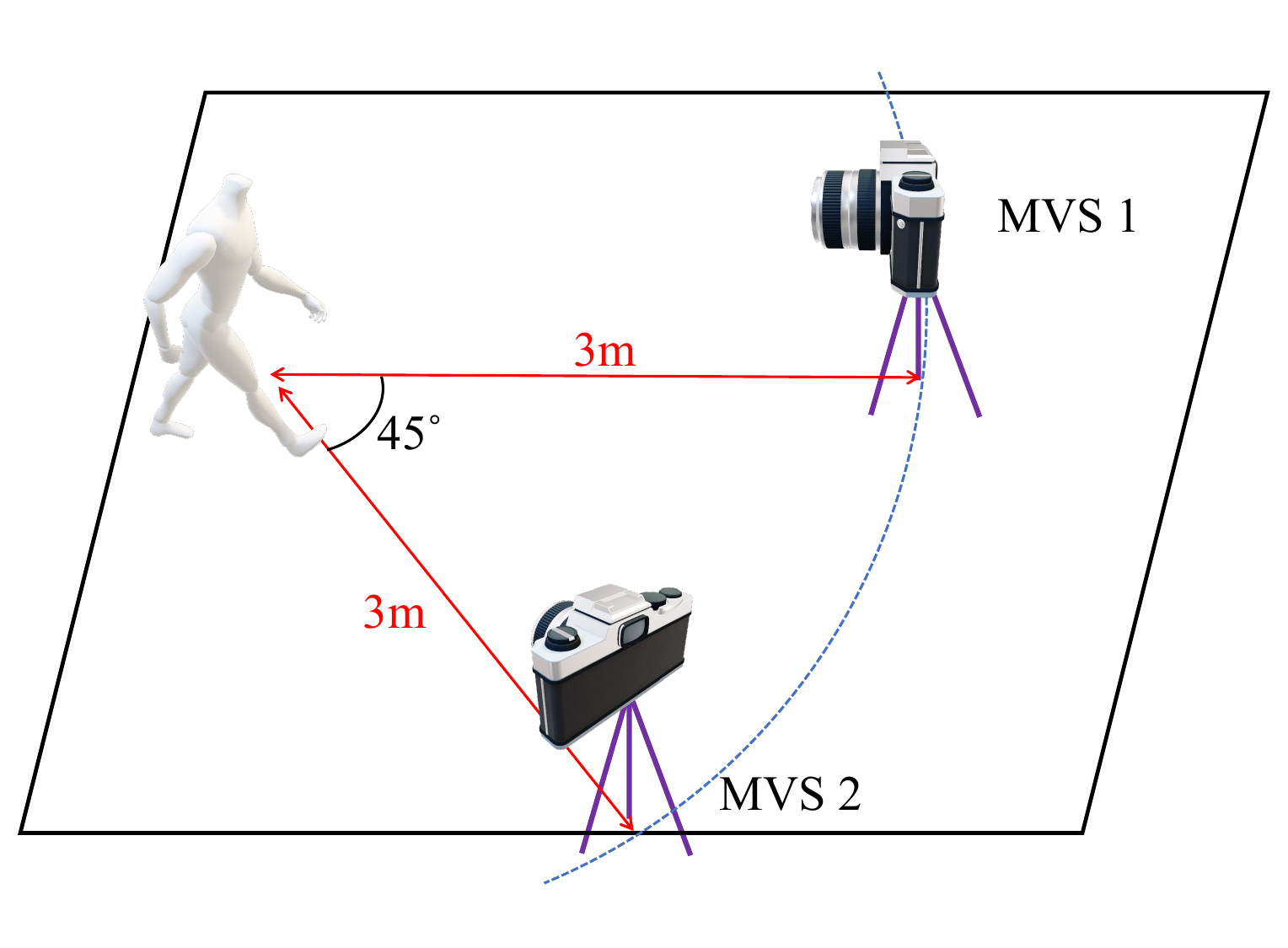}
        \vspace{-1.1\baselineskip}
        \caption{Environment Setting. In an area of 6$\times$5m$^{2}$, two MVSs are located 3 meters away from the subject, with a viewing angle of 45 degrees centered on the subject's position.}
        \label{fig:env}
    \end{minipage}
    \label{fig_fus_env}
    \vspace{-1.2\baselineskip}
     
\end{figure}

Edge and motion vector information complement each other. While edge is sufficient for detecting clear and non-blurred body joints, incorporating motion vectors into our model can effectively address the challenges posed by fast movements and overlapping boundaries, which may confuse edge-based HPE models. 
Hence, we aim to combine the complementary information of edge and motion vectors while keeping our model compact and efficient. We directly concatenate an edge image and a two-directional motion vector image, proposing the early fusion model (referred to as FS) as illustrated in Figure~\ref{fig:fusion}. Our FS model can leverage various single-branch network architectures designed for compactness and efficiency.

\subsection{Sparse Convolution}\label{sec.sparse-cov}
Sparse convolution methods, studied mainly on 3D point clouds~\cite{Wu_2019_CVPR} and 2D hand-written characters data~\cite{scn}, have shown on-par performance with dense methods while requiring substantially less computation. Our proposed edge and motion vector representation is 96.23\% and 99.13\% sparse on average, respectively. Hence, we leverage the submanifold sparse convolution network introduced in~\cite{scn} to train our compact fusion model. This approach offers a powerful tool for processing sparse data, allowing us to effectively exploit the sparsity of our representation while capturing the necessary information for accurate predictions. We also achieve on-par performance with 96\% fewer FLOPs compared to traditional convolution.

\vspace{-0.5\baselineskip}



\section{Our SPHP Dataset}

We use our proprietary MVS to collect an in-house edge and motion vector dataset with 16 types of actions performed by 40 users, as shown in Figure~\ref{fig:dataset}.

\noindent\textbf{Data Format.}
Motion Vector Sensor (MVS) captures three channels of 8-bit information, conveying edge, horizontal, and vertical motion vectors at each time frame. MVS has a resolution of 640$\times$480 and a field-of-view of 60 degrees. When collecting data, MVS records videos for 10 seconds per clip, with 30 frames per second.

\noindent\textbf{Environment Setting.} We collect the data in a room with an area of 6$\times$5m$^{2}$. Two MVSs are located 3 meters away from the subject, with a viewing angle of 45 degrees centered on the subject's position as shown in Figure~\ref{fig:env}. This setup allows us to collect data from two different viewing angles simultaneously, which increases the data collection efficiency.

\subsection{Participants and Actions}

We collected data from 40 subjects, having a balanced distribution of 20 male and 20 female participants. The average ages of male and female participants are 32 years and 22 years, respectively. Besides, the average heights of male and female participants are 172 and 161 centimeters, respectively. 
Participants performed 16 fitness-related actions, which are listed in Table~\ref{table:action_class} and categorized into four classes based on the type of movement: C1 for upper-body movements, C2 for lower-body movements, C3 for slow whole-body movements, and C4 for fast whole-body movements.

To diversify the viewing angles of our dataset, we apply a novel strategy to capture each action from multiple perspectives. Firstly, we place two cameras within an interval of 45 degrees. Then, we instruct the participants to face various directions (\textit{i.e.}, 0, 15, 30, and 45 degrees, respectively) while capturing their actions. In each direction, every participant will perform four actions, totaling 16 actions, as listed in Table~\ref{table:action_class}.
\begin{table}[htbp]
  \centering
  
  \caption{The 16 actions in our SPHP dataset are categorized into four classes: C1 for upper-body movements, C2 for lower-body movements, C3 for slow whole-body movements, and C4 for fast whole-body movements.}
    \resizebox{\columnwidth}{!}{\begin{tabular}{rrrl}
    \toprule    
    \multicolumn{1}{c}{C1} & \multicolumn{1}{c}{C2} & \multicolumn{1}{c}{C3} & \multicolumn{1}{c}{C4} \bigstrut\\
    \hline
    \multicolumn{1}{l}{1. Arm abduction} & \multicolumn{1}{l}{5. Leg knee lift} & \multicolumn{1}{l}{~~8. Squat} & 12. Elbow-to-knee \bigstrut[t]\\
    \multicolumn{1}{l}{2. Arm bicep curl} & \multicolumn{1}{l}{6. Leg abduction} & \multicolumn{1}{l}{~~9. Walk in place} & 13. Jump in place \\
    \multicolumn{1}{l}{3. Wave hello} & \multicolumn{1}{l}{7. Leg pulling} & \multicolumn{1}{l}{10. Standing side bend} & 14. Jumping jack \\
    \multicolumn{1}{l}{4. Punch up forward} &       & \multicolumn{1}{l}{11. Roll wrists \& ankles} & 15. Hop on one foot \\
          &       &       & 16. Jog in place \bigstrut[b]\\
    \bottomrule
    \end{tabular}}%
  \label{table:action_class}%
\end{table}%



\vspace{-0.8\baselineskip}

\subsection{Efficient Annotation}
Our dataset is annotated with the positions of 13 joints: nose, left/right shoulder, left/right elbow, left/right hand, left/right hip, left/right knee and left/right foot.
We use a well-trained grayscale HPE model to speed up the annotation process to mark the initial joint positions. Then, we manually confirm the initial position as acceptable or make fine-grained adjustments on each frame. All videos in our dataset are fully annotated.

\vspace{-0.5\baselineskip}

\subsection{Dataset Comparison}
We compare our \textbf{SPHP} dataset and two event-based datasets, namely DHP19~\cite{dhp19} and MMHPSD~\cite{eventhpe}, in Table~\ref{table:dataset_comp}.  Our dataset outperforms these datasets in various aspects. Firstly, our multi-modality data includes two-directional motion vectors, providing more detailed and nuanced information about actions. Secondly, SPHP boasts the highest number of subjects and frames, and exhibits greater diversity in different unique action types compared to DHP19, which has a large proportion of symmetric actions. Additionally, unlike DHP19 and MMHPSD, SPHP captures 10-second videos for each action rather than recording a fixed number of repetitions, thus providing more comprehensive information.


\begin{table}
  \begin{minipage}[c]{0.53\textwidth}
    \resizebox{\columnwidth}{!}{\begin{tabular}{c|ccccc}
    \hline
    Dataset & Sub\# & Seq\# & Frame\# & Dir.    & MM \bigstrut\\
    \hline
    DHP19~\cite{dhp19} & 17    & 33    & 87K   &     &  \bigstrut[t]\\
    MMHPSD~\cite{eventhpe} & 15    & 12    & 240K  &     & \checkmark \\
    SPHP (ours) & 40    & 16    & 384K  & \checkmark   & \checkmark \bigstrut[b]\\
    \hline
    \end{tabular}}%
  \end{minipage}\hfill
  \begin{minipage}[c]{0.45\textwidth}
    \caption{
       A comparison of our SPHP dataset with event-based datasets in terms of the number of subjects (Sub\#), action sequences (Seq\#) per subject, frames (Frame\#), directional motion (Dir), and multi-modality (MM).
    } \label{table:dataset_comp}
  \end{minipage}
  \vspace{-0.5\baselineskip}
\end{table}

    




\section{Experimental results}

\subsection{Human Pose Estimation}

To showcase the superiority of our sparse data in HPE, we conduct the experiments on three datasets: \textbf{SPHP}, MMHPSD~\cite{eventhpe} and HumanEva~\cite{humaneva}.
Within our SPHP dataset, we compare the performance across various input modalities, including edge, motion vectors, a fusion of edge and motion vectors, and grayscale images. Besides traditional convolution, we also employ submanifold sparse convolution to assess the computational efficiency gained from exploiting the sparsity in input data. Additionally, we assess the generalization capability of our method on MMHPSD and HumanEva. Notably, within the MMHPSD dataset, we incorporate the provided event data for further comparison with our motion vectors.

\noindent\textbf{Implementation.} 
We test on 3 CNN backbones, including the DHP19~\cite{dhp19} proposed model (218K), U-Net-Small (1.9M), and U-Net-Large (7.7M). U-Net-Small and U-Net-Large are built based on the architecture proposed by~\cite{unet}, incorporating three downsampling and upsampling operations. The output dimensions of the $3\times3$ convolutions in the U-Net-Large are twice as large as those of the U-Net-Small.
The input frames are resized to 288 $\times$ 384 for the SPHP dataset, whereas the MMHPSD dataset retains its original size of 256 $\times$ 256. For each joint, the model outputs a heatmap that indicates the likelihood of the joint position at each pixel. To generate a target heatmap for a joint, we initialize an all-zero map of the same size as the input frame and set a value of 1 to the pixel corresponding to the annotated joint position. The heatmap is then smoothed using Gaussian blur with $\sigma = 4$. Mean Squared Error (MSE) is employed as the loss function.

\noindent\textbf{Evaluation Metrics.} Mean Per Joint Position Error (\textbf{MPJPE}) is chosen for evaluation. ${\rm MPJPE} = \frac{1}{N}\sum_{i} \| y_{i}-\hat{y_{i}}\|$ calculates the Euclidean distance between predicted positions $\hat{y_{i}}$ and ground truth positions $y_{i}$ for each joint, where $N$ is the number of joints.



\begin{table}[htbp]
  \centering
  \caption{MPJPE (lower is better) on SPHP with different input types. Different input modalities, "GR", "ED", "MV", and "FS", are grayscale, edge, motion vector, and fusion (edge and motion vector), respectivetly. C1, C2, C3, C4 are four classes listed in Table~\ref{table:action_class}. C and SC are traditional and sparse convolutions, respectively.}
    \resizebox{\columnwidth}{!}{\begin{tabular}{cccrrrrrrrrrrr}
    \hline
    \multirow{2}[4]{*}{Backbone} & \multirow{2}[4]{*}{\# of Params} & \multirow{2}[4]{*}{Input} & \multicolumn{5}{c}{C}                 &       & \multicolumn{5}{c}{SC} \bigstrut\\
\cline{4-8}\cline{10-14}          &       &       & \multicolumn{1}{c}{C1} & \multicolumn{1}{c}{C2} & \multicolumn{1}{c}{C3} & \multicolumn{1}{c}{C4} & \multicolumn{1}{c}{Mean} &       & \multicolumn{1}{c}{C1} & \multicolumn{1}{c}{C2} & \multicolumn{1}{c}{C3} & \multicolumn{1}{c}{C4} & \multicolumn{1}{c}{Mean} \bigstrut\\
    \hline
    \multirow{4}[4]{*}{DHP19~\cite{dhp19}} & \multirow{4}[4]{*}{218K} & GR    & 2.62  & 3.08  & 3.33  & 3.62  & 3.20  &       & -     & -     & -     & -     & - \bigstrut\\
\cline{3-8}          &       & MV    & 16.96 & 6.50  & 6.43  & 5.11  & 8.66  &       & 56.96 & 27.12 & 25.86 & 14.12 & 30.20 \bigstrut[t]\\
          &       & ED    & 3.14  & 3.64  & 3.71  & 4.03  & 3.65  &       & 5.18  & 6.47  & 5.94  & 6.76  & 6.10 \\
          &       & FS    & 3.36  & 3.32  & 3.56  & 3.88  & \textbf{3.56} &       & 5.00  & 6.52  & 6.10  & 6.45  & \textbf{6.01} \bigstrut[b]\\
    \hline
    \multirow{4}[4]{*}{U-Net-Small} & \multirow{4}[4]{*}{1.9M} & GR    & 1.82  & 2.08  & 2.13  & 2.48  & 2.15  &       & -     & -     & -     & -     & - \bigstrut\\
\cline{3-8}          &       & MV    & 19.56 & 5.41  & 5.69  & 3.82  & 8.52  &       & 54.29 & 20.93 & 19.25 & 8.11  & 24.85 \bigstrut[t]\\
          &       & ED    & 3.20  & 3.49  & 3.19  & 3.49  & 3.35  &       & 3.35  & 3.78  & 3.48  & 3.95  & 3.65 \\
          &       & FS    & 3.32  & 2.91  & 2.79  & 3.18  & \textbf{3.07} &       & 3.42  & 3.61  & 3.41  & 3.69  & \textbf{3.54} \bigstrut[b]\\
    \hline
    \multirow{4}[4]{*}{U-Net-Large} & \multirow{4}[4]{*}{7.7M} & GR    & 1.73  & 2.14  & 2.00  & 2.33  & 2.06  &       & -     & -     & -     & -     & - \bigstrut\\
\cline{3-8}          &       & MV    & 18.76 & 5.27  & 5.54  & 3.79  & 8.25  &       & 52.08 & 20.00 & 18.65 & 7.77  & 23.86 \bigstrut[t]\\
          &       & ED    & 2.95  & 3.08  & 2.86  & 3.27  & 3.05  &       & 3.32  & 3.70  & 3.47  & 3.86  & 3.60 \\
          &       & FS    & 2.68  & 2.94  & 2.79  & 3.14  & \textbf{2.90} &       & 3.32  & 3.64  & 3.41  & 3.64  & \textbf{3.50} \bigstrut[b]\\
    \hline
    \end{tabular}}%
  \label{table:accuracy}%
  \vspace{-0.9\baselineskip}
\end{table}%


\begin{minipage}[t]{0.455\textwidth}
    \makeatletter\def\@captype{table}
    \caption{MPJPE at different speed levels. Each joint is classified into three levels, slow, medium, and fast. "ED," "MV," and "FS" stand for edge, motion vector, and fusion (edge and motion vectors).}

\label{tab:speed_levels}
\resizebox{\columnwidth}{!}{\begin{tabular}{ccccc}
    \toprule
    Backbone & Input & Slow  & Medium & Fast \bigstrut\\
    \hline
    \multirow{3}[2]{*}{DHP19~\cite{dhp19}} & MV    & 36.66 & 9.66  & 9.65 \bigstrut[t]\\
          & ED    & 5.66  & 6.90  & 8.22 \\
          & FS    & 5.69  & \textbf{6.76} & \textbf{7.34} \bigstrut[b]\\
    \hline
    \multirow{3}[2]{*}{U-Net-Small} & MV    & 31.07 & 5.33  & 4.77 \bigstrut[t]\\
          & ED    & 3.51  & 3.95  & 4.33 \\
          & FS    & \textbf{3.48} & \textbf{3.62} & \textbf{3.85} \bigstrut[b]\\
    \hline
    \multirow{3}[2]{*}{U-Net-Large} & MV    & 29.84 & 5.10  & 4.57 \bigstrut[t]\\
          & ED    & 3.47  & 3.82  & 4.22 \\
          & FS    & \textbf{3.44} & \textbf{3.62} & \textbf{3.84} \bigstrut[b]\\
    \bottomrule
    \vspace{0.8\baselineskip}
    \end{tabular}}%

\end{minipage}
\begin{minipage}[t]{0.02\textwidth}
    
\end{minipage}
\begin{minipage}[t]{0.455\textwidth}
    \makeatletter\def\@captype{table}

\caption{Comparison of GFLOPs and FPS (frames per second). All of the experiments are conducted for the fusion input modality on SPHP. "C" and "SC" are traditional and sparse convolutions.}
\label{table:sc}
\resizebox{\columnwidth}{!}{\begin{tabular}{ccccc}
\toprule
Backbone & Params & Conv. & GFLOPs & FPS \bigstrut\\
\hline
\multirow{3}[4]{*}{DHP19~\cite{dhp19}} & \multirow{3}[4]{*}{218K} & C     & 275.45 & 26.89 \bigstrut\\
\cline{3-5}      &       & \multirow{2}[2]{*}{SC} & \textbf{33.25} & \textbf{38.88} \bigstrut[t]\\
      &       &       & \textcolor[rgb]{ .122,  .122,  .122}{($\downarrow$87\%)} & \textcolor[rgb]{ .122,  .122,  .122}{(1.5x)} \bigstrut[b]\\
\hline
\multirow{3}[4]{*}{U-Net-Small} & \multirow{3}[4]{*}{1.9M} & C     & 1135  & 11.82 \bigstrut\\
\cline{3-5}      &       & \multirow{2}[2]{*}{SC} & \textbf{46.74} & \textbf{36.13} \bigstrut[t]\\
      &       &       & ($\downarrow$96\%)  & (3x) \bigstrut[b]\\
\hline
\multirow{3}[4]{*}{U-Net-Large} & \multirow{3}[4]{*}{7.7M} & C     & 4510  & 1.07 \bigstrut\\
\cline{3-5}      &       & \multirow{2}[2]{*}{SC} & \textbf{186.80} & \textbf{13.89} \bigstrut[t]\\
      &       &       & ($\downarrow$96\%)  & (13x) \bigstrut[b]\\
\bottomrule
\end{tabular}}%

\end{minipage}

\noindent\textbf{HPE Performance on SPHP.} 
The results in Table~\ref{table:accuracy} show that our fusion (FS) model achieves the best results compared to other sparse inputs (\textit{i.e.}, Edge (ED) and Motion Vector (MV)) for both traditional and sparse convolution networks. Besides, the gap between traditional and sparse convolutions is small in the U-Net-Small/U-Net-Large backbone, whereas the gap becomes more significant when using the smallest model (DHP19).

\begin{figure}[ht]
    \centering
    {\includegraphics[width = \linewidth]{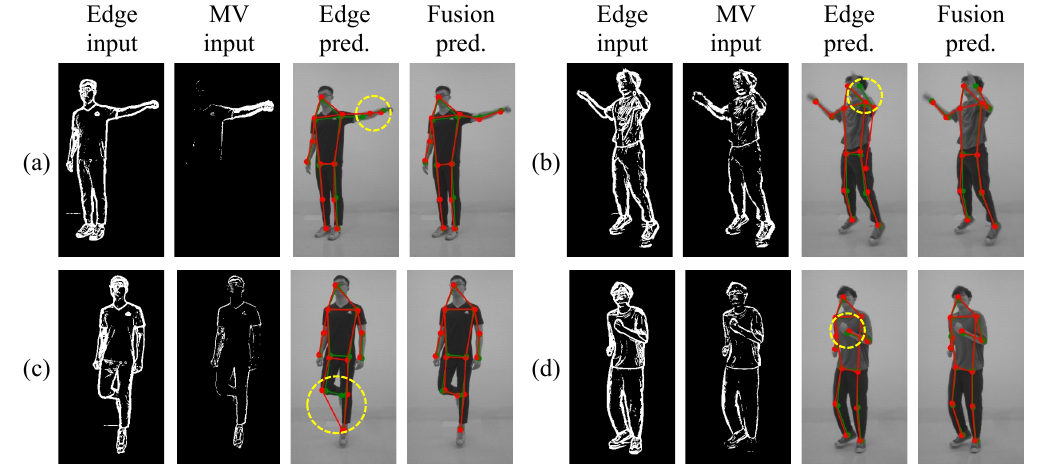}}
    \vspace{-0.5\baselineskip}
    \caption{Qualitative results. Our fusion (\textit{i.e.}, edge and MV)  method rectifies the inaccurately predicted joints caused by motion blur and unclear contour when using only edge inputs, as shown in yellow circles. Green dots denote the ground truth, and red ones represent the predictions. Note that the grayscale images in the figure are for better visualization only and are not used as input for both predictions.}
    \label{fig:improve}
    \vspace{-0.5\baselineskip}
\end{figure}

In Table~\ref{tab:speed_levels}, we evaluate MPJPE with sparse convolution networks based on the speed of each joint, categorized into three levels: slow, medium, and fast. For a 640$\times$480 image, a joint that moves less than 4 pixels compared to the previous frame is considered slow; between 4-6 pixels is medium; more than 6 pixels is fast. The results show that the gap between FS and ED is the largest for fast-moving joints. U-Net-Small shows the most noteworthy relative improvement of 11\% (from 4.33 to 3.85), validating that motion vectors complement edges in detecting fast movements. See supplementary materials for more performance analysis of backbones and joint speed.


The qualitative results are shown in Figure~\ref{fig:improve}. According to the results, the edge images alone can lead to motion blur and decreased accuracy. For instance, the joint predictions in Figure~\ref{fig:improve}(a)(b) are incorrect due to motion blur. Moreover, the unclear contour may also lead to inaccurate prediction, as shown in Figure~\ref{fig:improve}(c)(d). Performing early fusion on edge and motion data enables models to rectify pose estimation.

\noindent \textbf{Computational Efficiency.} As shown in Table~\ref{table:sc}, sparse convolution can achieve a noteworthy 96\% reduction in FLOPs.  We also evaluate frames per second (FPS) on an Intel Core i9-7940 3.1GHz CPU and observe a 3x and 13x acceleration in U-Nets. With sparse convolution, U-Net-Small strikes the best balance between accuracy (MPJPE) and speed (FPS).

\noindent\textbf{Experiments on more complex datasets.} 
We evaluate our method on the MMHPSD~\cite{eventhpe}, which provides event frames. On MMHPSD, we generate MV data using an off-the-shelf approach. The experimental results are presented in Table~\ref{tab:MMHPSD}. Notably, our "MV + edge" fusion approach exhibits superior performance compared to using edge or MV modalities alone. This shows the generalizability of our method across various datasets. Furthermore, using "MV" as input demonstrates superior performance compared to using the "event" provided by MMHPSD, both with and without edges. This serves as evidence that MV captures better information than traditional event cameras. Besides MMHPSD, we include the experimental results from HumanEva~\cite{humaneva} dataset in the technical report on our project page.

\begin{table}[htbp]
  \centering
  \caption{MPJPE on MMHPSD~\cite{eventhpe} with different input types. Conv. stands for convolutions.}
    \begin{tabular}{cccccc}
    \hline
          & Edge & Event & MV & Event + Edge & MV + Edge \bigstrut[t] \bigstrut[b]\\
    \hline
    Conv.     & 3.50  & 4.66  & 3.30  & 3.26  & \textbf{2.95} \bigstrut[t]\\
    Sparse Conv.    & 3.84  & 9.75  & 6.91  & 3.28  & \textbf{3.06} \bigstrut[b]\\
    \hline
    \end{tabular}%
    \vspace{-9pt}
  \label{tab:MMHPSD}%
\end{table}%

\subsection{Privacy-enhanced Representation}
\noindent\textbf{Face Recognition.}
To demonstrate the privacy-enhanced nature of our sparse data, we conduct face recognition experiments on CelebA~\cite{celebA} featuring over 10,000 individuals. We convert CelebA's RGB images into grayscale and edge images for comparison. Note that facial motion is typically small or absent, so it is not included. 
We perform face recognition on each type of image using the ResNet-50~\cite{resnet} backbone with Arcface~\cite{arcface} loss. 


Table~\ref{table:tabcelebA} shows that face recognition performances are similar for RGB and grayscale images. However, when using edge input, the accuracy drops by 4.1$\%$ compared to RGB, while the recall drops even more significantly by 10.2$\%$. This indicates that using the edge format can help reduce individuals' privacy exposure compared to the other two formats.


\begin{table}
  \begin{minipage}[c]{0.47\textwidth}
    \resizebox{\columnwidth}{!}{\begin{tabular}{c|cc|cc}
    \hline
    Input & Acc.  & Drop  & Recall  & Drop  \bigstrut\\
    \hline
    RGB   & 88.9  & -     & 84.1  & - \bigstrut[t]\\
    Grayscale  & 88.7  & 0.2   & 84.2  &  -0.1 \\
    Edge  & \textbf{84.8}  & 4.1   & \textbf{73.9}  & 10.2 \bigstrut[b]\\
    \hline
    \end{tabular}}%
  \end{minipage}\hfill
  \begin{minipage}[c]{0.5\textwidth}
    \caption{
      Face recognition performance for various input types on  CelebA~\cite{celebA}. We train the ResNet-50~\cite{resnet} backbone with Arcface~\cite{arcface} loss to perform face recognition. The recall of edge input drops by $10.2$.
    } \label{table:tabcelebA}
  \end{minipage}
  \vspace{-0.5\baselineskip}
\end{table}

\noindent\textbf{Human Ability for Cross-modality Face Matching.}
We further test the cross-modality face-matching ability of humans on our \textbf{SPHP} dataset. To simulate people identifying different faces, we design a survey to analyze the ability of 100 participants to match leaked edge faces with grayscale faces from different angles and appearances. Each participant will be asked ten questions. In each question, the participant should identify the person from ten choices of edge faces given a grayscale reference face. In order to make the test more complicated, faces in the choices are presented from various angles and differ from those in the questions.
The average test accuracy is only 18.8\%, whereas the accuracy increases to 87.3\% if we change the choices to grayscale images. These results suggest that people have a restricted ability to recognize daily faces from leaked edge images.

\section{Conclusion}
We introduce the novel \textbf{SPHP} dataset, which contains sparse edge images and two-directional motion vectors. Our proposed fusion model exhibits enhanced performance in human pose estimation compared to individual modalities using only edge or motion vectors. By leveraging our sparse data with submanifold sparse convolutions, we further reduce the FLOPs by 96\% and achieve a 13x speed-up compared to traditional convolutional neural networks. In addition, we showcase the generalization capability of our method through experiments on MMHPSD and HumanEva datasets. Furthermore, to verify the privacy-enhanced nature of our representation, we conduct a face recognition experiment using edge images as inputs, which results in a recall drop of 10.2\% compared to the use of RGB images. Finally, we carry out a user study to demonstrate the limited ability of humans to recognize faces across leaked edge and daily grayscale images.

\section{Acknowledgements}

This work is supported in part by Ministry of Science and Technology of Taiwan (NSTC 111-2634-F-002-022) and Novatek Microelectronics Corp.
We thank National Center for High-performance Computing (NCHC) for computational and storage resource. Besides, we appreciate Novatek's MVS technology in efficiently collecting edge and motion data for this project.

\bibliography{mybib/conference,mybib/others,mybib/hpe}
\end{document}